\documentclass{article} 
\usepackage[table]{xcolor}
\usepackage{iclr2027_conference,times}

\usepackage{amsmath,amsfonts,bm}

\def\eqref#1{equation~\ref{#1}}

\def\1{\bm{1}}

\DeclareMathAlphabet{\mathsfit}{\encodingdefault}{\sfdefault}{m}{sl}
\SetMathAlphabet{\mathsfit}{bold}{\encodingdefault}{\sfdefault}{bx}{n}

\usepackage{amsmath}
\usepackage{amssymb}
\usepackage{fvextra}
\usepackage{url}
\usepackage{booktabs}
\usepackage{multirow}
\usepackage{makecell}
\usepackage{graphicx}
\usepackage{array}
\usepackage{subcaption}
\usepackage{placeins}
\usepackage{tikz}
\usepackage{enumitem}
\usepackage{hyperref}
\usepackage{cleveref}
\newcommand{\commentout}[1]{}
\renewcommand{\paragraph}[1]{\noindent\textbf{#1.}}
\newcommand{\cmark}{\textcolor{green!50!black}{\ensuremath{\checkmark}}}
\newcommand{\xmark}{\textcolor{red!75!black}{\ensuremath{\times}}}

\title{ScanAR: Parallel Row-Scan Decoding for Efficient Image Generation in Unified Autoregressive Models}

\author{Jingen Qu \quad Xi Chen \quad Yingbo Tang \quad Xinghang Li$^{\dagger}$ \quad Tao Kong$^{\dagger}$ \\
Xiaomi Robotics \\
\texttt{\{qujingen,chenxi75,tangyingbo,lixinghang,kongtao\}@xiaomi.com}
}

\iclrfinalcopy
\begin{document}

\maketitle
\renewcommand{\thefootnote}{\fnsymbol{footnote}}
\footnotetext[2]{Corresponding authors: Tao Kong and Xinghang Li.}
\renewcommand{\thefootnote}{\arabic{footnote}}

\begin{abstract}
Unified multimodal autoregressive (AR) models represent text and images in one token sequence and generate all modalities with a shared next-token prediction head. Their image generation, however, decodes an $H\times W$ visual grid token by token in raster order, which takes $HW$ sequential steps and dominates inference time. Existing acceleration methods for visual AR generation either change the tokenizer, the prediction head, or the model interface, or keep the model but cannot reach high parallelism without losing quality, so they do not carry over to unified AR models.
We propose ScanAR, a row-scan parallel decoding method that only redefines the predecessors of visual tokens. The first visual row is generated from left to right as usual, and each later row is generated in a single parallel step, with every token predicted from the vertically aligned token of the previous row. Nonvisual tokens keep the standard autoregressive rule.
ScanAR reduces the visual decoding depth from $HW$ to $H+W$ without changing the backbone, vocabulary, or output head and without adding parameters, and it supports both training from scratch and lightweight post-training.
On Emu3.5-Image-34B, ScanAR post-training reaches a GenEval score of 82.36, the best among post-training acceleration methods trained on the same data, and generates $512\times512$ images $13.8\times$ faster than raster decoding under the same inference setting. On ImageNet, ScanAR trained from scratch has the highest throughput at every model size with FID comparable to raster LlamaGen. On the 34B Xiaomi-Robotics-U0 model, ScanAR preserves the overall quality of embodied scene generation, interleaved image--text generation, and video generation.
\end{abstract}

\section{Introduction}
\begin{figure}[!htbp]
    \centering
    \includegraphics[width=1\linewidth]{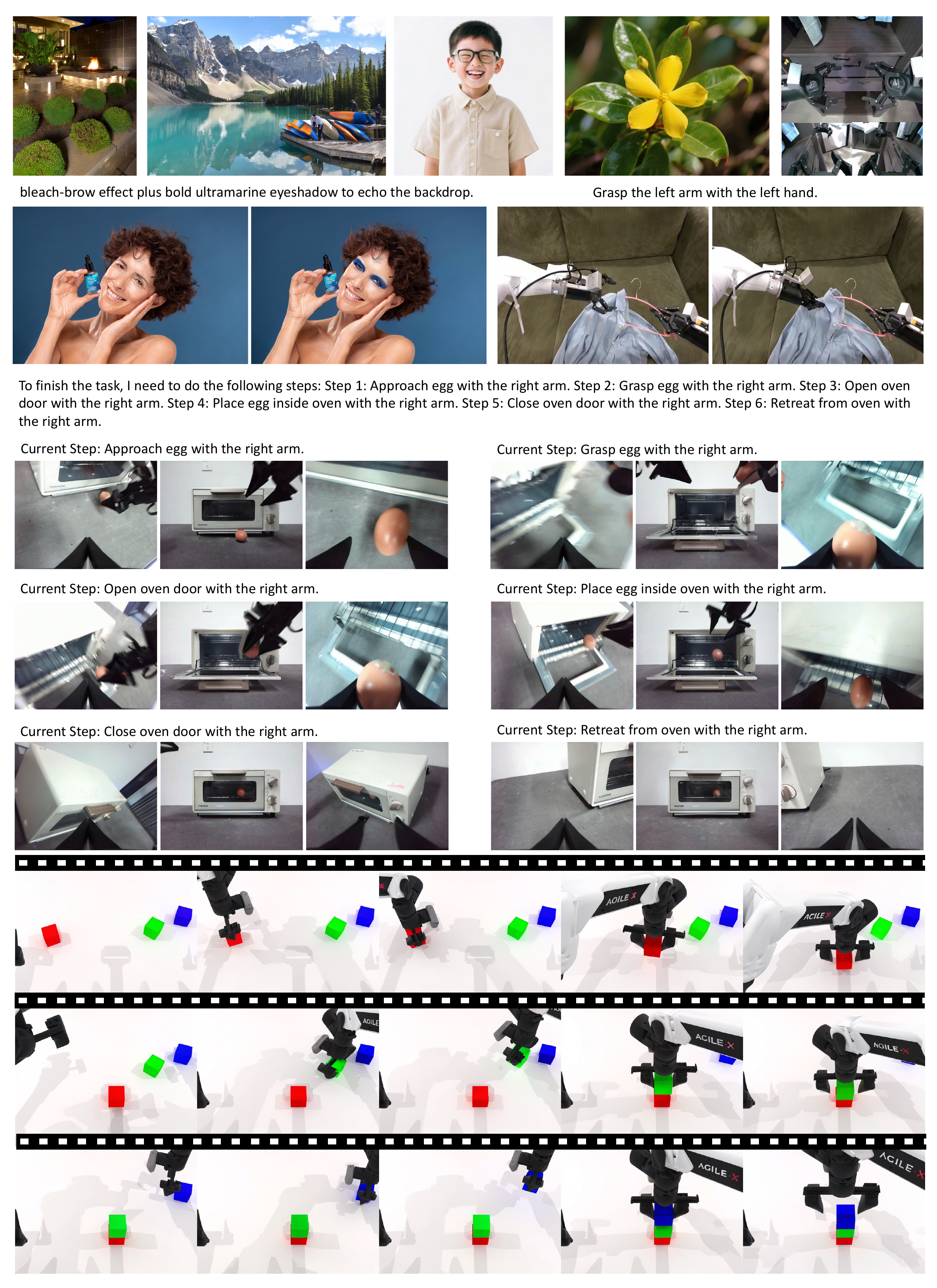}
    \caption{Qualitative results of ScanAR across diverse generation tasks, including text-to-image generation, image editing, interleaved image-text generation, and video generation.}
    \label{fig:all-img}
\end{figure}

The combination of discrete visual tokenizers and autoregressive Transformers casts image generation as a sequence modeling problem analogous to language modeling~\cite{esser2021taming,wang2024emu3}. Unified AR models further place textual and visual tokens in a shared vocabulary under a common next-token prediction interface, so that a single model can generate arbitrary mixed-modal sequences~\cite{chameleon2024,wang2024emu3,cui2025emu3}. This single interface is the main strength of the AR paradigm: text, images, and interleaved content come from one model with one decoding rule. Visual content, however, corresponds to a large two-dimensional token grid. Flattening this grid and decoding it token by token makes the visual spans dominate the sequential execution time of a unified generator. For an $H\times W$ image, standard raster AR requires $HW$ sequential decisions, which leads to high latency at practical resolutions.

Recent approaches to accelerating visual AR generation include scale-wise generation restructuring~\cite{tian2024var,han2025infinity}, spatial generation-path restructuring~\cite{wang2025par,li2026autoregressive,he2025nar,zhou2026flashar,zhang2026lpd}, and speculative decoding~\cite{teng2025sjd,wang2026csd}. Most of them are designed for a single-image canvas and change the tokenizer, the prediction head, or the model interface, which makes them hard to apply to an existing unified AR model. Methods that keep the model unchanged either remain causal within each row~\cite{he2025zipar}, which bounds their parallelism, or depend on the acceptance rate of drafted tokens~\cite{teng2025sjd}, which varies from image to image. An acceleration method for unified AR models should introduce visual parallelism inside the same conditional model that predicts text and image structure, keep its next-token interface intact, and give a fixed decoding depth for every image.

We propose ScanAR, an image-generation acceleration method that meets these requirements. ScanAR derives row-scan parallelism directly from the two-dimensional structure of image tokens, while all modalities remain modeled by a shared next-token prediction head. Text tokens, image-structure tokens, and tokens in the first visual row use their standard causal predecessors, whereas each subsequent visual token is predicted from the hidden state at the same column in the preceding row. This predecessor relation turns every later row into a single parallel prediction step, while rows still advance autoregressively. A row-block causal mask and a joint objective let the model learn standard causal and row-scan dependencies within the same prediction space, without any new parameters.

We evaluate ScanAR on a 34B unified model and on image-only benchmarks. On Xiaomi-Robotics-U0~\cite{li2026xiaomiroboticsu0}, ScanAR post-training preserves the overall quality of embodied scene generation, interleaved image--text generation, and video generation (Figure~\ref{fig:all-img}). On Emu3.5-Image-34B~\cite{cui2025emu3}, it reaches the best GenEval score among post-training acceleration methods trained on the same data and matches a raster model fine-tuned on that data, while generating $512\times512$ images $13.8\times$ faster than raster decoding under the same inference setting. On ImageNet, ScanAR trained from scratch achieves the highest throughput at every model size, and it reaches quality comparable to the training-free method ZipAR at its default setting with $4.5\times$ fewer forward passes. Our main contributions are as follows:

\begin{itemize}
    \item We introduce a two-dimensional row-scan decoding scheme that predicts each subsequent visual row in parallel, reducing the sequential decoding depth of an $H \times W$ visual grid from $HW$ to $H+W$.
    \item ScanAR keeps the raster token grid and row-major progression and adds no parameters, so it applies both to lightweight post-training of existing raster AR models and to training from scratch.
    \item ScanAR keeps the unified next-token interface. It accelerates the visual segments while preserving the ability to generate text, images, and interleaved sequences, which we verify quantitatively on interleaved and video generation.
\end{itemize}

\section{Related Work}
\label{related_work}

\subsection{Unified Autoregressive Models}

Unified AR models, exemplified by Chameleon and the Emu3 family, represent text and images as one sequence of discrete tokens and model it with a common autoregressive objective~\cite{chameleon2024,wang2024emu3,cui2025emu3}. A visual tokenizer encodes an image into a two-dimensional token grid, which is inserted into the sequence in a prescribed order. Textual and visual tokens share the same context, Transformer backbone, and output interface. This paradigm has a clear probabilistic formulation, inherits next-token prediction and scaling behavior from language modeling, and generates text, images, and interleaved multimodal content within a single model.

The main bottleneck of unified AR models lies in their image segments. A standard implementation flattens the visual grid into a raster sequence and generates it token by token. This preserves a strict causal factorization, but the inference depth of an image segment grows linearly with the number of visual tokens and far exceeds that of ordinary text generation at high resolutions.

\subsection{Acceleration of Visual Autoregressive Generation}

Existing approaches to visual AR acceleration fall into three groups: generation-granularity restructuring, spatial generation-path restructuring, and speculative decoding. Table~\ref{tab:related-comparison} summarizes them from the perspective of accelerating an existing unified raster AR model.

\paragraph{Generation-granularity restructuring}
These methods replace token-level prediction with scale-level or block-level prediction. VAR casts visual generation as coarse-to-fine next-scale prediction~\cite{tian2024var}, and Infinity scales this paradigm with a bitwise tokenizer and classifier~\cite{han2025infinity}. The number of sequential stages drops sharply, but the methods require a redesigned prediction objective and a multi-scale tokenizer. The learned distribution is no longer a next-token factorization over the original visual tokens, so they cannot be applied to an existing unified AR model.

\begin{table}[t]
    \centering
    \caption{Comparison of representative and recent approaches to visual AR inference acceleration. All columns refer to accelerating an existing unified raster AR model.}
    \label{tab:related-comparison}
    \small
    \setlength{\tabcolsep}{6pt}
    \renewcommand{\arraystretch}{1.12}
    \resizebox{\linewidth}{!}{%
\begin{tabular}{@{}lcccc@{}}
        \toprule
        \textbf{Method}
        & \shortstack{\textbf{Decoding}\\\textbf{depth}}
        & \shortstack{\textbf{No extra}\\\textbf{parameters}}
        & \shortstack{\textbf{Preserves raster}\\\textbf{grid and order}}
        & \shortstack{\textbf{Support}\\\textbf{unified output}} \\
        \midrule

        \multicolumn{5}{@{}l}{\textit{Baseline}} \\
        Unified raster AR~\cite{wang2024emu3} & $HW$ & \cmark & \cmark & \cmark \\

        \addlinespace[2pt]
        \multicolumn{5}{@{}l}{\textit{Generation-granularity restructuring}} \\
        VAR~\cite{tian2024var} & $10$ & \xmark & \xmark & \xmark \\
        Infinity~\cite{han2025infinity} & $13$ & \xmark & \xmark & \xmark \\
        DenseAR~\cite{park2026densear} & $64$ & \cmark & \xmark & \xmark \\

        \addlinespace[2pt]
        \multicolumn{5}{@{}l}{\textit{Spatial-path restructuring}} \\
        PAR~\cite{wang2025par} & $HW/p+p$ & \xmark & \xmark & \xmark \\
        ZipAR~\cite{he2025zipar} & $60$--$144$ & \cmark & \xmark & \cmark \\
        RandAR~\cite{pang2024randar} & $88$ & \xmark & \xmark & \xmark \\
        ARPG~\cite{li2026autoregressive} & $64$ & \xmark & \xmark & \xmark \\
        NAR~\cite{he2025nar} & $H+W-1$ & \xmark & \xmark & \xmark \\
        FlashAR~\cite{zhou2026flashar} & $H+W-1$ & \xmark & \xmark & \xmark \\
        LPD~\cite{zhang2026lpd} & $20/48$ & \xmark & \xmark & \xmark \\
        RadAR~\cite{wang2025radar} & $10/13$ & \xmark & \xmark & \xmark \\

        \addlinespace[2pt]
        \multicolumn{5}{@{}l}{\textit{Speculative/Jacobi decoding}} \\
        SJD~\cite{teng2025sjd} & Adaptive & \cmark & \cmark & \xmark \\
        PJD~\cite{liao2026pjd} & Adaptive & \cmark & \xmark & \xmark \\
        CSD~\cite{wang2026csd} & Adaptive & \cmark & \cmark & \xmark \\
        SSD~\cite{xiang2026ssd} & $n$ (square) & \xmark & \xmark & \xmark \\

        \addlinespace[2pt]
        \multicolumn{5}{@{}l}{\textit{Ours}} \\
        \textbf{ScanAR (ours)} & $H+W$ & \cmark & \cmark & \cmark \\
        \bottomrule
    \end{tabular}%
}

    \parbox{0.98\linewidth}{\vspace{0.45em}\scriptsize
    \cmark/\xmark\ denotes yes/no.
    \textit{No extra parameters}: no parameters are added on top of the raster AR model being accelerated, and its checkpoint is reused.
    \textit{Preserves raster grid and order}: tokens are produced on the same $H\times W$ grid, and each row is completed before the next row starts.
    \textit{Support unified output}: the shared next-token interface over text and image tokens is kept.
    Numeric depths are reported configurations (RandAR uses 256 visual tokens; Infinity uses 13 scales for $1024^2$; ZipAR uses windows $w=2$ to $8$ on a $16\times16$ grid, measured with the official code); symbolic depths omit constant offsets. Here, $p$ is the PAR parallel width and $n$ the side length of a square grid.}
\end{table}

\paragraph{Spatial generation-path restructuring}
These methods reorganize the raster decoding topology to exploit the two-dimensional structure of visual tokens. Representative approaches include token grouping in PAR, neighborhood expansion in NAR, random-order decoding in RandAR and ARPG, learnable position queries in LPD, and directional prediction in FlashAR~\cite{wang2025par,he2025nar,pang2024randar,li2026autoregressive,zhang2026lpd,zhou2026flashar}. They reach low decoding depths, but most of them add prediction parameters or change the input and prediction interface, and they are trained as dedicated image generators. ZipAR~\cite{he2025zipar} is a training-free alternative that starts decoding a row once the previous row has produced a fixed window of tokens. It keeps the original model, but it retains strict left-to-right causality inside every row, so its depth stays above the row length and grows with the window. As we show in Section~\ref{sec:exp-zipar}, its quality degrades quickly as the window shrinks.

\paragraph{Speculative decoding}
Speculative decoding drafts multiple candidate tokens and verifies them with the target model~\cite{leviathan2023fast}. SJD extends this idea to visual generation with speculative Jacobi decoding~\cite{teng2025sjd}, and later work improves the drafting and verification steps~\cite{liao2026pjd,wang2026csd,xiang2026ssd}. Their speedup depends on draft quality, acceptance rate, and verification cost, which vary across images and modalities.

\paragraph{Positioning of ScanAR}
ScanAR uses a deterministic row-frontier dependency. It keeps the raster grid and row-major progression, predicts each row in parallel from the completed previous row, and shares one output head for standard causal and row-scan prediction. Applied independently to each visual segment, it supports dynamic image sizes and general unified AR outputs. It is the only method in Table~\ref{tab:related-comparison} that reaches $H+W$ depth with no extra parameters while preserving unified output.

\section{Method}
\label{methods}

\subsection{Unified Prediction Space}

Given a multimodal prefix $x$, a unified autoregressive (AR) model generates a sequence $z$ containing text, control tokens, and a variable number of image blocks. Image boundaries and spatial dimensions are generated as structural tokens. ScanAR accelerates each visual segment through row-scan prediction while retaining the original backbone, vocabulary, and shared next-token prediction head.

Text and structural tokens retain their standard causal predecessors. Visual tokens use the predecessor mapping defined below, independently for each image block. This formulation supports arbitrary image positions and dimensions within a unified output stream.

\subsection{Row-Scan Prediction}

For an $H\times W$ visual grid, let $y_{i,j}$ denote the token at row $i$ and column $j$, using zero-based indices. ScanAR assigns each token to the generation step
\begin{equation}
s(i,j)=
\begin{cases}
j, & i=0,\\
W+i-1, & i>0.
\end{cases}
\label{eq:step-id}
\end{equation}
The first row is generated from left to right, and each later row forms one parallel group. 
Equation~\eqref{eq:step-id} counts $H+W-1$ sampling steps. For $H>1$, one additional state-only first-row commit (described below) gives an effective visual decoding depth of $H+W$, compared with $HW$ for raster AR; a single-row grid uses $W$ steps.

Let $h_{\mathrm{pref}}^{\mathrm{causal}}$ denote the causal hidden state of the image prefix. This state is used to predict the first visual token. For the first row, let $h_{0,j}^{\mathrm{causal}}$ denote the hidden state of visual token
$y_{0,j}$ after it has been committed under causal attention. Let
$\bar{h}_{i,j}$ denote the hidden state at position $(i,j)$ after the
complete row $i$ has been submitted with full visibility within that
row.

The prediction state for visual token $y_{i,j}$ is defined as

\begin{equation}
g_{i,j} =
\begin{cases}
h_{\mathrm{pref}}^{\mathrm{causal}}, & i=0,\ j=0,\\
h_{0,j-1}^{\mathrm{causal}}, & i=0,\ j>0,\\
\bar{h}_{i-1,j}, & i>0.
\end{cases}
\label{eq:predecessor}
\end{equation}

Thus, the tokens in the first row are predicted using causal hidden states.
Starting from the second row, the token at column $j$ uses the fully visible hidden state
of the token at the same column in the preceding row.

Since the first row is generated causally, the hidden state of each first-row token only incorporates the preceding tokens. 
These causal hidden states cannot directly serve as the predecessors for the second row, whose tokens should condition on the complete first row.

After the first row has been generated, ScanAR performs one
non-sampling first-row commit operation. The generated first-row tokens
are reprocessed together with the image prefix, while all tokens within
the first row are made mutually visible. This operation produces the
fully visible states $\bar{h}_{0,j}$, which are then used to predict the
second row.

No separate commit operation is required for later rows. From the second
row onward, all tokens in the current row are submitted to the backbone
as a group. During this forward pass, every query in the current row can
attend to all tokens in that row, in addition to the prefix and all
previously completed rows. Therefore, the resulting hidden states are
already the fully visible states $\bar{h}_{i,j}$ and can be used directly
as the predecessors for row $i+1$.

Accordingly, the decoding schedule consists of $W$ causal generation
steps for the first row, one non-sampling commit step, and one parallel
generation step for each subsequent row. The generated tokens within a
parallel row are conditionally independent given the image prefix and
all completed preceding rows, while the row-level autoregressive
dependency is preserved. Compared with raster AR, the context of token $y_{i,j}$ differs only by the tokens to its left in the same row, at most $W-1$ tokens. All previously generated rows, which carry the global layout of the image, remain fully visible.




\begin{figure}[t]
    \centering

    \begin{minipage}[c]{0.30\linewidth}
        \centering
        \begin{subfigure}[c]{\linewidth}
            \centering
            \includegraphics[
                width=\linewidth,
                height=0.42\textheight,
                keepaspectratio
            ]{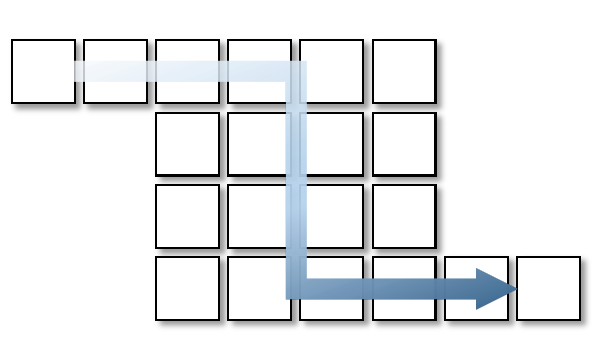}
            \caption{ScanAR.}
            \label{fig:result-e}
        \end{subfigure}
    \end{minipage}
    \hspace{0.018\linewidth}
    \begin{minipage}[c]{0.012\linewidth}
        \centering
        \begin{tikzpicture}
            \draw[
                dashed,
                line width=1.4pt,
                dash pattern=on 5pt off 4pt
            ] (0,0.1) -- (0,0.3\textheight);
        \end{tikzpicture}
    \end{minipage}
    \hspace{0.018\linewidth}
    \begin{minipage}[c]{0.62\linewidth}
        \centering

        \begin{subfigure}[t]{0.48\linewidth}
            \centering
            \includegraphics[width=\linewidth]{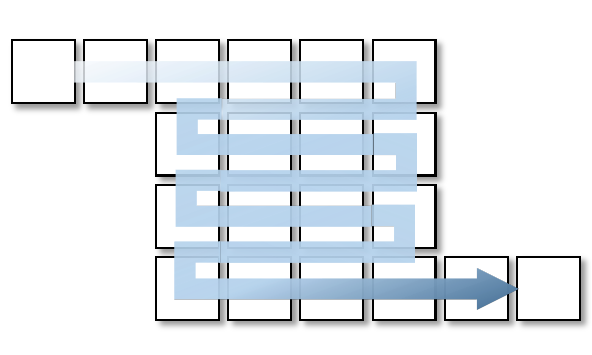}
            \caption{Raster AR.}
            \label{fig:result-a}
        \end{subfigure}
        \hfill
        \begin{subfigure}[t]{0.48\linewidth}
            \centering
            \includegraphics[width=\linewidth]{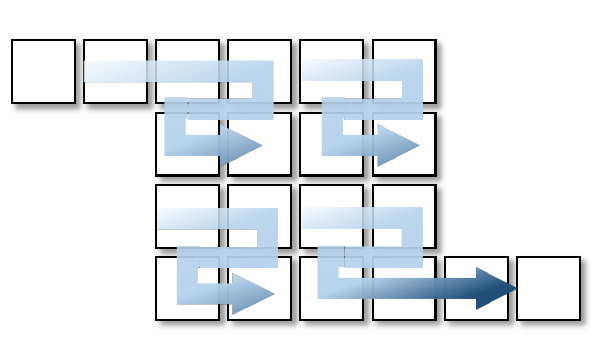}
            \caption{PAR.}
            \label{fig:result-b}
        \end{subfigure}

        \vspace{0.8em}

        \begin{subfigure}[t]{0.48\linewidth}
            \centering
            \includegraphics[width=\linewidth]{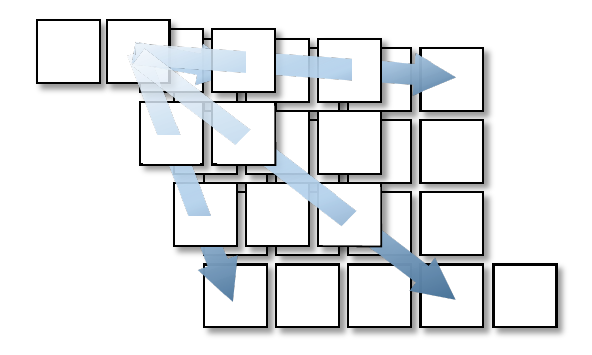}
            \caption{VAR.}
            \label{fig:result-c}
        \end{subfigure}
        \hfill
        \begin{subfigure}[t]{0.48\linewidth}
            \centering
            \includegraphics[width=\linewidth]{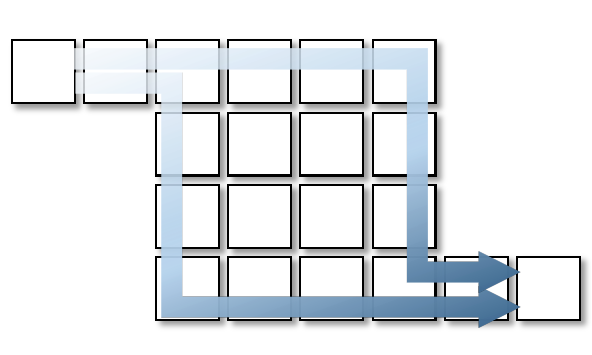}
            \caption{FlashAR.}
            \label{fig:result-d}
        \end{subfigure}
    \end{minipage}

    \caption{Comparison of five different AR decoding methods. ScanAR uses a concise row-scan design that remains compatible with multimodal AR generation.}
    \label{fig:five-panel-layout}
\end{figure}

\subsection{Attention Mask and Training}

The mask specifies hidden-state context, and Equation~\eqref{eq:predecessor} selects the state used to predict each target. Let $R_i$ denote the positions of visual row $i$ and $b_i$ its first position. For queries and keys in the original sequence, the visibility set is
\begin{equation}
    \mathcal{V}(q)=
    \begin{cases}
        \{k\mid k\le q\}, & q\text{ is nonvisual or }q\in R_0,\\
        \{k\mid k<b_i\}\cup R_i, & q\in R_i,\ i>0.
    \end{cases}
    \label{eq:hybrid-mask}
\end{equation}
Nonvisual positions and the first row use causal attention. Later rows use full within-row attention to form committed states for subsequent predictions. We refer to this mask as the \emph{row-block causal mask}: attention is causal across rows and full within each later row.

\begin{figure}[h]
    \centering
    \includegraphics[width=0.5\linewidth]{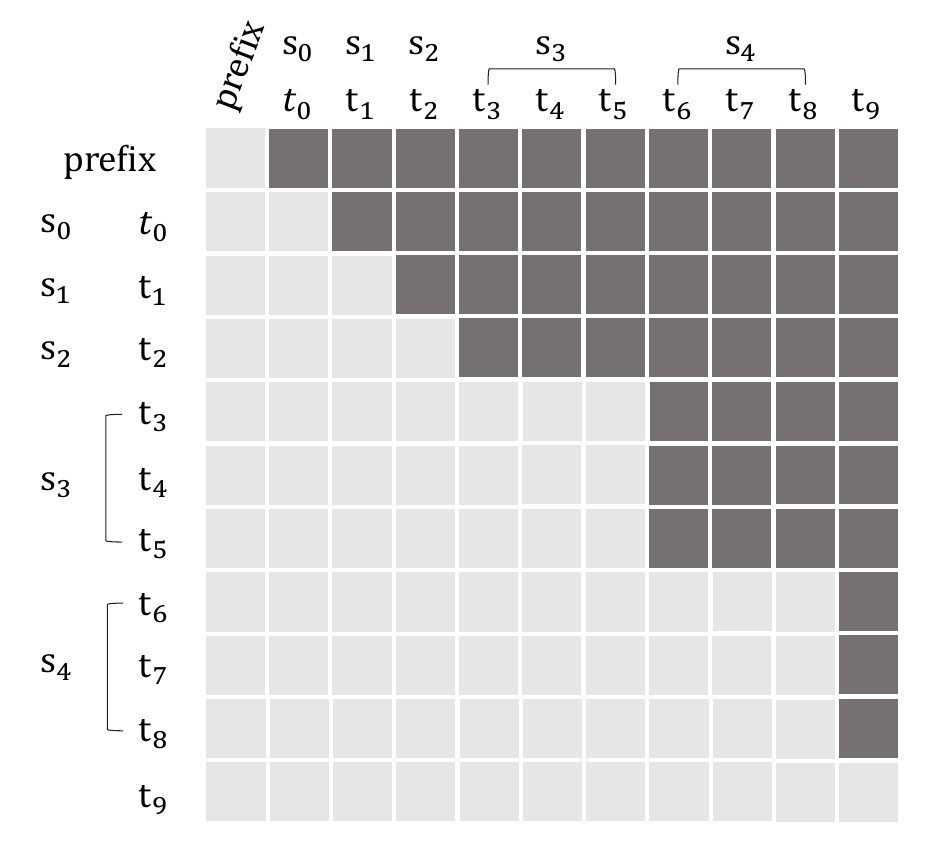}
    \caption{Training visibility and predecessor mapping of ScanAR over a unified sequence.}
    \label{fig:training-mask}
\end{figure}

Training uses ground-truth tokens throughout. An auxiliary copy of the first row, with identical token and position IDs, computes $\bar h_0$ in the same forward pass as the original sequence. Its queries attend only to the preceding context and their own complete copy; original sequence positions attend only to original positions. The copy has no independent prediction targets and receives gradients through the second-row loss. Figure~\ref{fig:training-mask} illustrates the mask and predecessor mapping.

For each image, visual cross-entropy is evaluated on the logits obtained from the selected predecessor states:
\begin{equation}
    \mathcal{L}_{\mathrm{vis}}
    =-\frac{1}{HW}\sum_{i=0}^{H-1}\sum_{j=0}^{W-1}
    \log\operatorname{softmax}(\ell_{i,j})_{y_{i,j}}.
    \label{eq:visual-loss}
\end{equation}
Text and structural targets use standard preceding-position cross-entropy. The joint objective optionally includes visual distribution alignment with a frozen raster-AR teacher~\cite{hinton2015distilling}:
\begin{equation}
    \mathcal{L}
    =\mathcal{L}_{\mathrm{vis}}
    +\lambda_{\mathrm{txt}}\mathcal{L}_{\mathrm{txt/ctrl}}
    +\lambda_{\mathrm{d}}(t)T^2
    \operatorname{KL}\!\left(p_{\mathrm{AR}}^{(T)}\,\|\,p_{\mathrm{Scan}}^{(T)}\right),
    \label{eq:joint-loss}
\end{equation}
where the KL term is averaged over aligned visual targets at temperature $T$. Both models receive the ground-truth sequence: the teacher uses raster-causal predecessors, and the student uses $g_{i,j}$. During post-training, $\lambda_{\mathrm{d}}(t)$ is annealed to zero after initial alignment; it is zero when training from scratch. ScanAR supports both regimes without adding trainable parameters.

\subsection{Unified Inference and KV Cache}

Inference evaluates the same predecessor mapping using sampled tokens. First-row tokens are sampled and appended to the cache individually. The first-row commit then recomputes all $W$ positions through every decoder layer with full within-row attention, reusing the preceding context's KV. The resulting $\bar h_0$ is stored separately for second-row prediction; temporary commit KV is discarded, and the main cache retains the original causal first-row KV.

For each later row, ScanAR samples all tokens from the preceding row's committed states, then processes the sampled row through every decoder layer with full within-row attention. This forward pass produces its committed states and appends its layer-wise KV at the original sequence positions. Earlier cache entries remain fixed.

After the image, text and structural tokens resume standard causal decoding over the accumulated cache, which contains the causal first-row KV and the subsequent rows' KV computed with within-row attention. The separate first-row commit states supply prediction sources without adding output tokens or replacing the sequence cache.

\section{Experiments}

\subsection{Experimental Setup}

We evaluate ScanAR in two settings. The primary setting uses the 34B Xiaomi-Robotics-U0 unified multimodal autoregressive model~\cite{li2026xiaomiroboticsu0}. The secondary setting follows the image-only protocols of prior work and covers class-conditional and text-to-image generation. Results from the two settings are reported separately because they use different base models, datasets, and evaluation protocols.

For the unified multimodal setting, ScanAR is initialized from a multimodal Xiaomi-Robotics-U0 checkpoint and retains the original visual tokenizer, unified vocabulary, Transformer backbone, and output head. We perform full-parameter post-training without adding trainable parameters. The training corpus contains image-generation and interleaved multimodal data, including image editing, subject preservation, multi-reference conditioning, spatial control, embodied-scene generation and transfer, video generation, and interleaved image-text generation. Training uses visual-token cross-entropy together with distribution-level supervision from a frozen raster-AR teacher during the initial adaptation stage. Detailed settings are given in Appendix~\ref{app:unified-model-details}.

For direct comparison with prior work, the image-only experiments follow the protocol of FlashAR~\cite{zhou2026flashar}. They evaluate LlamaGen~\cite{sun2024llamagen} on class-conditional ImageNet generation~\cite{russakovsky2015imagenet}\footnote{This paper uses the ImageNet dataset (\url{https://huggingface.co/datasets/ILSVRC/imagenet-1k}). The authors confirm that the use of this dataset in this paper is solely for academic research purposes and not for any commercial activities.} and Emu3.5-Image-34B~\cite{cui2025emu3} on text-to-image synthesis. Their protocols are described in Appendix~\ref{app:image-only-details}.

\subsection{Unified Multimodal Generation}
\label{sec:exp-unified}

We test whether row-scan decoding preserves the capabilities of the unified model on three tasks: embodied-scene transfer, interleaved image--text generation, and video generation.

\paragraph{Embodied-scene transfer}
This task tests conditional scene generation under depth and multi-view constraints. Outputs are multi-view canvases of about one megapixel (about 4{,}096 visual tokens, the token count of a $1024^2$ image). We use GPT-5.4 as an automated evaluator. Given the scene instruction, depth maps, and generated multi-view images, it scores six dimensions on a 10-point scale (Appendix~\ref{app:gpt54-evaluation}). Both models are evaluated on the same 300 prompts (150 single-dimension and 150 multi-dimension edits from five embodied sources) with the same judge and prompt template.

As shown in Table~\ref{tab:u0-embodied-transfer}, ScanAR obtains an overall score of $7.49$, compared with $7.52$ for the original model. The two models are thus on par overall. ScanAR is slightly higher on depth faithfulness and multi-view consistency, the two dimensions that most directly test conditional geometric control. Per-sample statistics are given in Appendix~\ref{app:u0-stats}.

\begin{table}[h]
\centering
\scriptsize
\setlength{\tabcolsep}{3pt}
\caption{GPT-5.4 evaluation of embodied-scene transfer on U0 (300 paired prompts). Scores are on a 10-point scale; higher is better.}
\label{tab:u0-embodied-transfer}
\resizebox{\linewidth}{!}{%
\begin{tabular}{lrrrrrrr}
\toprule
Method
& \shortstack{Depth\\Faithfulness}
& Background
& Foreground
& Lighting
& \shortstack{Image\\Quality}
& \shortstack{Multi-View\\Consistency}
& Overall \\
\midrule
U0     & 7.4 & 7.6 & 7.3 & 7.5 & 8.0 & 7.5 & 7.52 \\
ScanAR & 7.5 & 7.4 & 7.2 & 7.5 & 7.8 & 7.6 & 7.49 \\
\bottomrule
\end{tabular}%
}
\end{table}

\paragraph{Interleaved image--text generation}
We compare the ScanAR-adapted U0 with the official raster-AR Xiaomi-Robotics-U0-Sequence model on 80 embodied subtask-planning cases from nine robot data sources. Each case provides a task instruction and initial observations, and the model must write the subtask plan and generate the observation images for each step (13.8 images per case on average). Both models run in the same inference engine on the same cases, so the image-generation path differs only in raster versus row-scan visual decoding. The metrics were fixed before generation and are all reported in Table~\ref{tab:interleave}; their definitions are given in Appendix~\ref{app:interleave-metrics}.

\begin{table}[h]
\centering
\small
\caption{Paired evaluation of interleaved image--text generation on 80 embodied cases.}
\label{tab:interleave}
\begin{tabular}{lrr}
\toprule
Metric & Raster U0-Seq & ScanAR \\
\midrule
Valid output (finished, all images decoded) $\uparrow$ & 81.3\% & 91.3\% \\
Number of steps $=$ requested $\uparrow$ & 100\% & 98.8\% \\
Correct text/image alternation per step $\uparrow$ & 100\% & 98.8\% \\
Image fidelity to GT, DINOv2 cosine $\uparrow$ & 0.670 & 0.654 \\
Image fidelity to GT, LPIPS $\downarrow$ & 0.397 & 0.408 \\
Image fidelity to GT, PSNR $\uparrow$ & 15.13 & 15.06 \\
Scene consistency with initial observation $\uparrow$ & 0.616 & 0.577 \\
Temporal consistency between adjacent steps $\uparrow$ & 0.690 & 0.672 \\
Subtask-text exact match with GT $\uparrow$ & 53.6\% & 80.1\% \\
Generation time per image $\downarrow$ & 57.3 s & \textbf{5.2 s} ($11.1\times$) \\
\bottomrule
\end{tabular}
\end{table}

ScanAR keeps the structure of the interleaved output: it finishes the sequence and alternates text and images as reliably as the raster model. Fidelity of the generated images to the ground-truth observations is close to that of the raster model on all three metrics. Similarity to the initial observation and between adjacent steps is slightly lower. The full interleaved sequence, including text, is generated $11.1\times$ faster end to end. Since ScanAR does not change the text path, we attribute the difference in subtask-text accuracy to the post-training data rather than to the decoding scheme.

\paragraph{Video generation}
We evaluate video generation on the public WorldArena RobotWin 2.0 test set (1{,}000 episodes) with its official 11-metric protocol, covering visual quality, temporal consistency, motion, and VLM-judged task metrics. We compare ScanAR with a raster U0 video model from the training run that provides ScanAR's initialization, on the same episodes and prompts; each model uses its default sampling configuration. Results are shown in Table~\ref{tab:worldarena}.

\begin{table}[h]
\centering
\small
\caption{WorldArena RobotWin 2.0 test set (1{,}000 episodes). Each $28\times37$-token frame is decoded in 65 sequential steps by ScanAR instead of 1{,}036 by raster AR.}
\label{tab:worldarena}
\begin{tabular}{llrr}
\toprule
Group & Metric & Raster U0 & ScanAR \\
\midrule
\multirow{2}{*}{Visual quality} & Aesthetic Quality & 41.50 & 41.00 \\
 & Image Quality & 54.16 & 53.16 \\
\midrule
\multirow{4}{*}{Temporal consistency} & Subject Consistency & 79.08 & 77.32 \\
 & Background Consistency & 86.21 & 84.95 \\
 & Motion Smoothness & 95.24 & 94.43 \\
 & Photometric Consistency & 1.76 & 2.60 \\
\midrule
\multirow{2}{*}{Motion} & Dynamic Degree & 73.07 & 67.58 \\
 & Flow Score & 82.38 & 74.97 \\
\midrule
\multirow{3}{*}{VLM-judged task} & Instruction Following & 80.20 & \textbf{83.12} \\
 & Interaction Quality & 74.28 & \textbf{77.40} \\
 & Perspectivity & 94.62 & \textbf{96.32} \\
\midrule
 & \textbf{Overall} & 69.32 & 68.44 \\
\bottomrule
\end{tabular}
\end{table}

ScanAR stays within 0.9 points of the raster model in overall score while decoding each frame in 16$\times$ fewer sequential steps. It scores higher on all three VLM-judged task metrics, which assess whether the robot carries out the instructed task, and its visual-quality and consistency metrics are within about 1--2 points. The main difference is a lower amount of motion (Dynamic Degree and Flow Score), which measures how much the video moves rather than whether the motion is correct.

\paragraph{Summary}
Across the three tasks, the per-metric changes go in both directions, while the overall quality is preserved and the task-level metrics improve. Because the ScanAR models are post-trained on a narrow embodied mixture, part of these per-metric changes may come from post-training rather than from decoding; we discuss this in Section~\ref{sec:limitations}. The controlled comparison on Emu3.5 below isolates the effect of decoding.

\subsection{Image-Only Benchmarks}

\paragraph{Class-conditional generation}
To show that ScanAR can also be learned from scratch, we train LlamaGen-based ScanAR models at four scales on ImageNet $256\times256$. Table~\ref{tab:llamagen-short} reports FID, IS, precision/recall, decoding steps, and throughput. ScanAR and raster LlamaGen use the same data and number of epochs. With $8\times$ fewer sequential steps, ScanAR and LlamaGen each obtain the better FID at some model sizes (ScanAR is better at XL), and the differences are small. ScanAR has the highest IS and throughput at every model size: its IS increases from 244.3 for B to 306.9 for XXL, and its throughput is $5.7\times$ (B) to $4.6\times$ (XXL) that of LlamaGen. Methods with lower FID, such as NAR and FlashAR, add prediction parameters and do not support unified output (Table~\ref{tab:related-comparison}); FlashAR is additionally post-trained from a converged raster model. Additional details are provided in Appendix~\ref{LlamaGen}.

\begin{table*}[h]
\centering
\small
\caption{Quantitative evaluation on the ImageNet $256\times256$ benchmark.}
\label{tab:llamagen-short}

\resizebox{\linewidth}{!}{%
\begin{tabular}{c l l c c c c c c}
\toprule
Model size & Method & Type & Epoch & FID$\downarrow$ & IS$\uparrow$
& P/R-F1$\uparrow$ & Steps & \shortstack{Throughput\\(img/s)}
\\
\midrule

\multirow{7}{*}{\shortstack{B\\($\sim120$M)}}
& LlamaGen & From scratch & 300 & 5.46 & 193.6 & 0.594 & 256 & 117.9 \\
& PAR & From scratch & 300 & 6.21 & 204.4 & 0.537 & 67 & 174.1 \\
& NAR & From scratch & 300 & \textbf{4.65} & 212.3 & 0.600 & 31 & 419.7 \\
& BlockDiffusion & Post-training & 75 & 5.91 & 176.2 & 0.589 & 64 & 186.3 \\
& FlashAR & Post-training & 25 & 4.68 & 208.3 & \textbf{0.605} & 31 & 447.2 \\
& ZipAR & Training-free & -- & 5.69 & 192.9 & 0.568 & 144 & 158.2 \\
& ScanAR & From scratch & 300 & 5.59 & \textbf{244.3} & 0.568 & 32 & \textbf{669.4} \\
\midrule

\multirow{8}{*}{\shortstack{L\\($\sim360$M)}}
& LlamaGen & From scratch & 300 & 3.80 & 248.3 & 0.639 & 256 & 47.1 \\
& PAR & From scratch & 300 & 4.32 & 189.4 & 0.576 & 67 & 93.8 \\
& NAR & From scratch & 300 & \textbf{3.06} & 263.9 & 0.641 & 31 & 195.4 \\
& VAR & From scratch & 200 & 3.30 & 274.4 & 0.634 & 10 & 129.3 \\
& BlockDiffusion & Post-training & 75 & 4.55 & 243.5 & 0.645 & 64 & 103.2 \\
& FlashAR & Post-training & 25 & 3.16 & 289.0 & \textbf{0.656} & 31 & 224.7 \\
& ZipAR & Training-free & -- & 3.69 & 226.6 & 0.636 & 144 & 60.8 \\
& ScanAR & From scratch & 300 & 3.86 & \textbf{295.2} & 0.637 & 32 & \textbf{235.2} \\
\midrule

\multirow{6}{*}{\shortstack{XL\\($\sim700$M)}}
& LlamaGen & From scratch & 300 & 3.39 & 227.1 & 0.648 & 256 & 23.7 \\
& PAR & From scratch & 300 & 3.50 & 234.4 & 0.619 & 67 & 53.9 \\
& NAR & From scratch & 300 & \textbf{2.70} & 277.5 & \textbf{0.676} & 31 & 98.1 \\
& BlockDiffusion & Post-training & 75 & 4.13 & 258.6 & 0.654 & 64 & 41.7 \\
& FlashAR & Post-training & 25 & 2.94 & 293.7 & 0.672 & 31 & 109.3 \\
& ScanAR & From scratch & 300 & 3.34 & \textbf{301.8} & 0.648 & 32 & \textbf{113.4} \\
\midrule

\multirow{6}{*}{\shortstack{XXL\\($\sim1.4$B)}}
& LlamaGen & From scratch & 300 & 3.09 & 253.6 & 0.647 & 256 & 14.1 \\
& PAR & From scratch & 300 & 3.20 & 288.3 & 0.632 & 67 & 33.9 \\
& NAR & From scratch & 300 & \textbf{2.58} & 293.5 & 0.673 & 31 & 56.9 \\
& BlockDiffusion & Post-training & 75 & 3.78 & 264.9 & 0.652 & 64 & 26.8 \\
& FlashAR & Post-training & 25 & 2.79 & 289.4 & \textbf{0.690} & 31 & 63.4 \\
& ScanAR & From scratch & 300 & 3.46 & \textbf{306.9} & 0.640 & 32 & \textbf{65.3} \\
\bottomrule
\end{tabular}%
}
\end{table*}

\paragraph{Comparison with training-free parallel decoding}
\label{sec:exp-zipar}
ZipAR~\cite{he2025zipar} accelerates a raster model without training by starting each row once the previous row has produced $w$ tokens. Table~\ref{tab:llamagen-short} includes its best setting ($w=8$, its default) on the same raster LlamaGen checkpoints. ZipAR reaches a quality comparable to ScanAR but needs 144 forward passes, $4.5\times$ ScanAR's 32, and ScanAR is $4.2\times$ (B) and $3.9\times$ (L) faster. With smaller windows, ZipAR's quality collapses (Appendix~\ref{app:zipar}). ZipAR asks a raster-trained model to predict tokens whose left context is not yet available, whereas ScanAR is trained on exactly the row-scan conditionals it uses at inference time; this training is what makes high parallelism possible without losing quality.

\paragraph{Text-to-image generation}
We post-train Emu3.5-Image-34B with ScanAR under the FlashAR protocol, in which BlockDiffusion, FlashAR, and ScanAR use the same 80M-token post-training data. Table~\ref{tab:emu-summary} reports post-training cost, latency, and GenEval. ScanAR uses 20K post-training steps, compared with 50K for the other methods, and has the lowest latency and the highest GenEval score among them. It reduces the serial visual decoding depth from 1024 to 64. With the two CFG branches batched in both models, the latency drops from 64.60 to 4.69 seconds, a $13.8\times$ speedup. Section~\ref{sec:efficiency} analyzes this speedup.

\begin{table}[h]
\centering
\small
\caption{Training cost, inference latency, and overall GenEval at $512\times512$ resolution. All methods batch the conditional and unconditional CFG branches (batch size 2).}
\label{tab:emu-summary}
\begin{tabular}{l l r r r r}
\toprule
Method & Type & Steps & Data & Latency (s)$\downarrow$ & GenEval$\uparrow$ \\
\midrule
Emu3.5-Image & From scratch & 940K & 150B & 64.60 & 80.48\\
BlockDiffusion & Post-training & 50K & 80M & 6.17 & 73.83\\
FlashAR & Post-training & 50K & 80M & 5.68 & 80.29\\
\textbf{ScanAR} & \textbf{Post-training} & \textbf{20K} & \textbf{80M}
& \textbf{4.69} & \textbf{82.36} \\
\bottomrule
\end{tabular}
\end{table}

The detailed GenEval results are reported in Table~\ref{tab:geneval-emu}. To separate the effect of the post-training data from that of decoding, we also fine-tune Emu3.5-Image with standard raster AR on the same data, with the same initialization, batch size, and learning-rate schedule (2K steps, five seeds). This raster model reaches $82.01 \pm 0.85$, so the gain over the original model mainly comes from the post-training data. ScanAR keeps this quality while decoding in $16\times$ fewer sequential steps, whereas BlockDiffusion and FlashAR fall below even the original model. ScanAR is also on par with or above the raster fine-tune on the two layout-level relational categories, Counting ($69.06$ vs.\ $66.50$) and Position ($80.00$ vs.\ $77.95$). A per-category analysis, including a direct test of within-row relations, is given in Appendix~\ref{app:geneval-analysis}.

\begin{table}[h]
\centering
\small
\caption{GenEval scores on Emu3.5-Image-34B at $512\times512$ resolution. The raster fine-tune uses the same post-training data and is reported as mean over five seeds.}
\label{tab:geneval-emu}
\resizebox{\linewidth}{!}{%
\begin{tabular}{lrrrrrrr}
\toprule
Method & Overall & Single Obj. & Two Obj. & Counting & Colors
& Position & Color Attr. \\
\midrule
Emu3.5-Image & 80.48 & \textbf{100.00} & \textbf{94.95} & 53.75 & 90.96 & 73.00 & \textbf{70.25} \\
Raster fine-tune (reference) & 82.01 & 99.19 & 92.22 & 66.50 & 88.88 & 77.95 & 67.30 \\
\midrule
BlockDiffusion & 73.83 & 96.88 & 88.89 & 47.50 & 85.64 & 68.00 & 58.44 \\
FlashAR & 80.29 & 98.75 & 91.92 & 53.75 & \textbf{92.55} & 80.00 & 64.00 \\
\textbf{ScanAR} & \textbf{82.36} & 99.06 & 87.88 & \textbf{69.06} & 90.16 & \textbf{80.00} & 68.00 \\
\bottomrule
\end{tabular}%
}
\end{table}

\subsection{Training and Inference Efficiency}
\label{sec:efficiency}

ScanAR introduces no trainable parameters and retains all AR model components. Its inference schedule uses only single-token steps and full-row steps, which reduces the visual decoding depth while keeping a regular execution pattern.

\paragraph{Training}
With a FlexAttention implementation of the row-block causal mask, ScanAR keeps training efficiency close to that of the causal AR baseline. At a sequence length of 8K, a ScanAR training step takes 21.3 seconds, compared with 20.2 seconds for causal AR. Detailed measurements are given in Appendix~\ref{app:efficiency}.

\paragraph{Speedup under matched inference settings}
To measure the algorithmic speedup on its own, we run raster Emu3.5-Image in the same inference code as ScanAR, with the same batched CFG, KV cache, attention kernels, GPU, prompts, and sampler. The raster decoder reproduces the reference next-token distributions up to bf16 precision. Raster decoding needs 1{,}055 sequential forward passes per $512\times512$ image (1{,}024 visual tokens and 31 end-of-row tokens) and takes 64.60 seconds; ScanAR needs 64 and takes 4.69 seconds. The $13.8\times$ speedup is slightly below the $16.5\times$ reduction in forward passes, because a ScanAR row step processes 32 query tokens instead of one (about 74 vs.\ 61 ms per forward pass). Decoding cost is dominated by reading the weights and the KV cache, so a full-row step costs only slightly more than a single-token step. Measurement details are given in Appendix~\ref{app:matched-latency}.

\paragraph{Serving}
We further evaluate vLLM-based implementations on a single H20E GPU at $512\times512$ resolution. After warm-up, 64 requests are submitted concurrently, and the reported time per image is derived from aggregate throughput, so it represents amortized generation time rather than single-request latency (Table~\ref{tab:scanar_inference_efficiency}). Compared with FlashAR, ScanAR reduces the amortized time per image from 1.37 to 1.15 seconds, increases throughput by 19.1\%, and lowers peak memory by 28.1 GB. The regular row-wise schedule makes execution predictable and reuses standard serving kernels efficiently.

\begin{table}[h]
\centering
\caption{Inference efficiency of vLLM-based implementations on a single H20E GPU at $512\times512$ resolution. After warm-up, 64 requests are submitted concurrently. Time per image is derived from aggregate throughput across all requests.}
\label{tab:scanar_inference_efficiency}
\small
\begin{tabular}{@{}lrrr@{}}
\toprule
Method & Time per image (s) $\downarrow$ & Throughput (images/s) $\uparrow$ & Peak memory (GB) $\downarrow$ \\
\midrule
FlashAR & 1.37 & 0.73 & 138.8 \\
ScanAR  & 1.15 & 0.87 & 110.7 \\
\bottomrule
\end{tabular}
\end{table}

\section{Limitations}
\label{sec:limitations}

Tokens within a parallel row are sampled conditionally independently given the completed rows, so ScanAR approximates the raster factorization rather than reproducing it exactly. The approximation removes at most $W-1$ tokens of context per token and is learned during training, but it may matter more for content with strong fine-grained dependencies along a row. The first row is still generated token by token, which leaves $W$ sequential steps per image.

Our U0 comparisons are not fully controlled. The raster references differ from the ScanAR models in checkpoint and post-training, and the video comparison uses each model's default sampling configuration. The ScanAR models are post-trained on a narrow mixture of embodied subtask plans and robot videos, which is a possible reason for the per-metric changes in Section~\ref{sec:exp-unified}, such as the lower amount of video motion and the lower similarity to the first observation. A raster U0 model post-trained on the same data and settings would separate these effects, but it requires retraining a 34B model on 32K-token video sequences, and we leave it to future work. On Emu3.5-Image, where such a control is feasible, row-scan decoding matches raster fine-tuning on the same data.

\section{Conclusions}
\label{conclusions}
We presented ScanAR, a row-scan parallel autoregressive method that accelerates image generation in unified AR models without modifying the backbone, vocabulary, or output head and without adding parameters. ScanAR keeps the shared token sequence and next-token prediction objective, leaves the standard autoregressive dependencies unchanged for nonvisual tokens and the first visual row, and predicts every later row in parallel from the vertically aligned tokens of the preceding row. This reduces the sequential decoding depth of an $H \times W$ image from $HW$ to $H+W$ and applies independently to each image segment of an interleaved multimodal sequence. ScanAR supports both training from scratch and lightweight post-training of existing unified AR models. On Emu3.5-Image-34B, it achieves the best GenEval score among post-training acceleration methods and a $13.8\times$ speedup over raster decoding under matched inference settings. On ImageNet, it achieves the highest throughput at every model size. On the 34B U0 model, it preserves the overall quality of embodied scene generation, interleaved generation, and video generation. These results show that ScanAR is a simple and practical way to make image generation efficient within unified AR models.

\bibliography{iclr2027_conference}
\bibliographystyle{iclr2027_conference}

\appendix

\section{Detailed Experimental Settings}

\subsection{Unified Model Setting}
\label{app:unified-model-details}

The Xiaomi-Robotics-U0 model comprises 64 decoder-only Transformer layers with
a hidden dimension of 5,120, an FFN dimension of 25,600, 64 attention heads,
and 8 grouped-query attention groups.

We optimize ScanAR for 10K steps with a global batch size of 512 using Adam
with $\beta_1=0.9$, $\beta_2=0.95$, $\epsilon=10^{-8}$, a weight decay of 0.1,
and gradient clipping at 5.0. The learning rate is linearly warmed up to
$1\times10^{-5}$ during the first $1\%$ of training and then decayed to
$1\times10^{-6}$ using a cosine schedule.

During the first 1K steps, a frozen raster-scan autoregressive teacher provides
distribution-level supervision at all target visual positions with a
distillation temperature of 2.0. The distillation weight is linearly annealed
from 0.3 to 0, while the visual-token cross-entropy objective remains active
throughout training. Training uses bfloat16 precision with FP32 attention
softmax and gradient reduction.

\subsection{Image-Only Model Setting}
\label{app:image-only-details}

We follow exactly the experimental protocol of FlashAR
\cite{zhou2026flashar}. These experiments are conducted independently from the
U0/ScanAR experiments.

We evaluate LlamaGen~\cite{sun2024llamagen} for class-conditional image
generation on ImageNet~\cite{russakovsky2015imagenet} at
$256\times256$ resolution, and Emu3.5-Image-34B~\cite{cui2025emu3} for
text-to-image synthesis at $512\times512$ resolution.

For LlamaGen, post-training is conducted for 25 epochs on ImageNet with a batch
size of 512; training from scratch is conducted for 300 epochs on ImageNet with a batch size of 512. For Emu3.5-Image-34B, we use approximately 80K text-image pairs
curated from OpenGPT-4o-Image and ShareGPT-4o-Image
\cite{chen2025opengpt4oimage,chen2025sharegpt4oimage}, and perform
post-training for 20K steps.

The training uses a two-stage schedule. During the first $20\%$ of the training steps, we incorporate a distillation loss. During the second stage, the distillation loss is removed. Both stages use a base learning rate of $2\times10^{-5}$ with cosine decay. The auxiliary-loss coefficient is set to $\lambda_{\mathrm{aux}}=0.05$. The classifier-free guidance scale is set to $2.0$ for LlamaGen and $5.0$ for Emu3.5-Image-34B. All experiments are performed on eight NVIDIA H200 GPUs using bfloat16 precision.

We compare against the original raster-scan autoregressive model, parallel
autoregressive models trained from scratch, including VAR
\cite{tian2024var}, PAR \cite{wang2025par}, and NAR
\cite{he2025nar}, as well as the post-training adaptation method
BlockDiffusion~\cite{cui2025blockdiffusion}.

For LlamaGen, we report FID
\cite{heusel2017fid}, Inception Score (IS)
\cite{salimans2016inception}, precision, recall, and sFID. For
Emu3.5-Image-34B, we use GenEval
\cite{ghosh2023geneval} to measure compositional fidelity. Inference latency
is measured on a single NVIDIA H20-141G GPU with one image per request (the two CFG branches of the accelerated methods are batched) and averaged
over 100 generated samples.

\section{More Experimental Results}

\subsection{GPT-5.4 Evaluation Protocol}
\label{app:gpt54-evaluation}

We use GPT-5.4 to evaluate depth-conditioned multi-view image generation in
embodied robot-manipulation scenes. For each evaluation sample, the evaluator
receives a textual scene description, a set of depth maps, and the
corresponding generated images. The $i$-th generated image is paired with the
$i$-th depth map. All views depict the same scene from different camera
angles.

The depth maps encode normalized inverse depth within each individual view:
brighter pixels indicate regions closer to the camera, while darker pixels
indicate regions farther away. Because the values represent relative depth
rather than absolute scale, depth faithfulness is evaluated independently for
each view.

GPT-5.4 assigns a score from 1 to 10 for each of the following six dimensions:

\begin{enumerate}
    \item \textbf{Depth Faithfulness}: whether the generated image follows the
    spatial layout, depth ordering, occlusion relationships, and geometric
    structure specified by the corresponding depth map.

    \item \textbf{Background}: whether background objects, walls, distant
    furniture, and other far-field elements agree with the scene description,
    while avoiding hallucinated background content.

    \item \textbf{Foreground}: whether the robot arm, gripper, manipulated
    objects, and other near-field elements are present and correctly arranged
    according to the scene description.

    \item \textbf{Lighting}: whether illumination direction, intensity, color
    temperature, shadows, highlights, and reflections are plausible and
    consistent across views.

    \item \textbf{Image Quality}: whether the generated images are clear,
    sharp, realistic, and free from artifacts such as blur, noise, color
    banding, distorted geometry, unnatural materials, or broken object
    boundaries.

    \item \textbf{Multi-View Consistency}: whether objects, poses, colors,
    textures, lighting, and scene geometry remain consistent across camera
    views.
\end{enumerate}

The scores follow a common qualitative scale: scores of 9--10 indicate
near-perfect results with no noticeable flaws; scores of 7--8 indicate good
results with only minor inaccuracies; scores of 5--6 indicate acceptable
results with clear but non-critical errors; scores of 3--4 indicate substantial
failures; and scores of 1--2 indicate severely degraded or unusable results.

The evaluator applies two special rules. First, if the numbers of depth maps
and generated images differ, only the pairs that can be matched in order are
evaluated, and the mismatch itself is not penalized. Second, when only one
view is available, multi-view consistency is not applicable and receives a
score of 10.

The evaluator returns a JSON object containing one score and one textual
explanation for each dimension, together with an overall score. The overall
score reflects the overall quality of the generated results and is not required
to be the strict arithmetic mean of the six dimension scores. The complete
evaluation prompt is provided below.

\begin{quote}
\begin{Verbatim}[
  breaklines=true,
  breaksymbolleft={},
  breaksymbolright={},
  fontsize=\scriptsize
]
You are an expert evaluator for depth-conditioned multi-view image generation in robot manipulation scenes.

The task being evaluated is: given a set of depth maps (one per camera view) and a text description of the scene, generate photorealistic images that faithfully follow the depth layout and match the text description.

You are given two sets of images:

1. {num_views} depth map(s), each representing the normalized inverse depth from a different camera view of the same scene. In these maps, brighter pixels are closer to the camera and darker pixels are farther away. The values represent relative depth within each individual image, not absolute scale.
2. {num_views} generated image(s), each showing the model's output for the corresponding camera view.

The depth maps and generated images are provided in the same order - the i-th generated image corresponds to the i-th depth map.

All views together depict the same scene from different camera angles.

---------------------
Evaluation Dimensions
---------------------

Score each dimension from 1 to 10.

General scoring guideline:

9--10: Excellent - the generated image is near-perfect in this dimension. No noticeable flaws; any imperfection is negligible and would not be caught without careful side-by-side comparison.
7--8: Good - overall successful with only minor issues. A few small inaccuracies exist but do not significantly detract from the result. Most viewers would consider it high quality.
5--6: Acceptable - the general intent is captured but there are clear, easily noticeable problems. Some elements are wrong, misplaced, or missing, though the image is still recognizable as an attempt at the target.
3--4: Poor - significant failures in this dimension. Multiple major errors are present; the result only partially resembles what was expected.
1--2: Very poor - almost entirely incorrect or broken. The image fails to satisfy the basic requirements of this dimension.

Evaluate the following dimensions:

1. Depth Faithfulness
Does each generated image faithfully follow the spatial layout defined by its corresponding depth map?

The depth map is a normalized inverse depth image: brighter = closer, darker = farther. It only encodes relative depth within a single view and does not represent absolute scale. Evaluate each view independently - do NOT compare depth consistency across different views.

Special rule: if the number of depth maps and generated images differ, only evaluate the pairs that can be matched in order (e.g., if there are 3 depth maps and 2 generated images, only evaluate the first 2 pairs). Do NOT penalize for the mismatch in count.

Check (per view):
- Whether near and far objects are placed at correct relative distances as indicated by the depth map
- Whether the depth ordering of objects is preserved (closer objects properly occlude farther ones)
- Whether the overall 3D spatial structure (surfaces, furniture, workspace layout) matches the depth information
- Whether depth gradients and transitions (e.g., a table surface receding, walls at different distances) are correctly reflected
- Whether the robot arm, gripper, and manipulated objects respect the depth relationships shown in the map

2. Text-Image Alignment - Background / Far Field
Do the generated images accurately reflect the background and distant elements described in the scene description?

Important: since multiple views together describe one scene, an object mentioned in the text only needs to appear in at least one view to count as present. Not every view needs to show every described element.

Check:
- Whether background objects, walls, distant furniture, and far-field elements mentioned in the text are present in at least one view
- Whether the overall room layout and environmental context match the description
- Whether there are hallucinated far-field objects not described in the text

3. Text-Image Alignment - Foreground / Near Field
Do the generated images accurately reflect the foreground and near-field elements described in the scene description?

Important: since multiple views together describe one scene, an object mentioned in the text only needs to appear in at least one view to count as present. Not every view needs to show every described element.

Check:
- Whether foreground objects, the robot arm, gripper, manipulated objects, and nearby items mentioned in the text are present in at least one view
- Whether spatial relationships among near-field objects (e.g., "on top of", "next to", "inside") are correctly rendered where visible
- Whether there are hallucinated near-field objects not described in the text

4. Text-Image Alignment - Lighting
Do the generated images accurately reflect the lighting conditions described or implied by the scene description?

Check:
- Whether the overall illumination direction, intensity, and color temperature match the described or expected lighting
- Whether shadows are consistent with the light source direction and scene geometry
- Whether highlights, reflections, and ambient light behave naturally
- Whether lighting is consistent across views (same light source direction and intensity)

5. Image Quality
Is each generated image visually clear, sharp, realistic, and free of artifacts?

Check:
- Whether there are visible artifacts such as blurring, noise, color banding, or distortions
- Whether colors and exposure are natural and well-balanced
- Whether there are structural distortions (warped edges, broken geometry, misshapen surfaces)
- Whether material appearances (reflection, transparency, roughness) look natural and realistic
- Whether fine-grained textures are clearly rendered (wood grain, metal sheen, fabric weave, etc.)
- Whether object edges are crisp and clean without jagged or smeared boundaries

6. Multi-View Consistency
Are the generated images geometrically and visually consistent with each other?

Special rule: if there is only 1 depth map or only 1 generated image, multi-view consistency is not applicable - score this dimension 10 automatically.

All views depict the same scene from different angles. They should be coherent as a set.

Check:
- Whether the same object appears at geometrically consistent positions across views
- Whether colors, textures, and lighting conditions are uniform across views
- Whether there are contradictions (e.g., an object present in one view but missing in another, or appearing with different shapes or colors)
- Whether the robot arm and gripper pose is consistent across all views
- Whether the overall scene layout is coherent when considering all views together

---------------------
Output Requirements
---------------------

Return JSON only.

Do not include any text outside the JSON.

{
  "depth_faithfulness": {"score": 0, "explanation": ""},
  "text_image_alignment_background": {"score": 0, "explanation": ""},
  "text_image_alignment_foreground": {"score": 0, "explanation": ""},
  "text_image_alignment_lighting": {"score": 0, "explanation": ""},
  "image_quality": {"score": 0, "explanation": ""},
  "multi_view_consistency": {"score": 0, "explanation": ""},
  "overall_score": 0
}

The overall_score should reflect the overall quality of the generation considering all dimensions (not necessarily a strict average).

Below is the text description of the scene, followed by the depth maps:

Instruction:
{instruction}

Here are the depth maps (one per camera view):

<image1><image2><image3>
\end{Verbatim}
\end{quote}

\subsection{U0 Result}
\label{app:u0}

\begin{figure}[!htbp]
    \centering
    \includegraphics[width=1\linewidth]{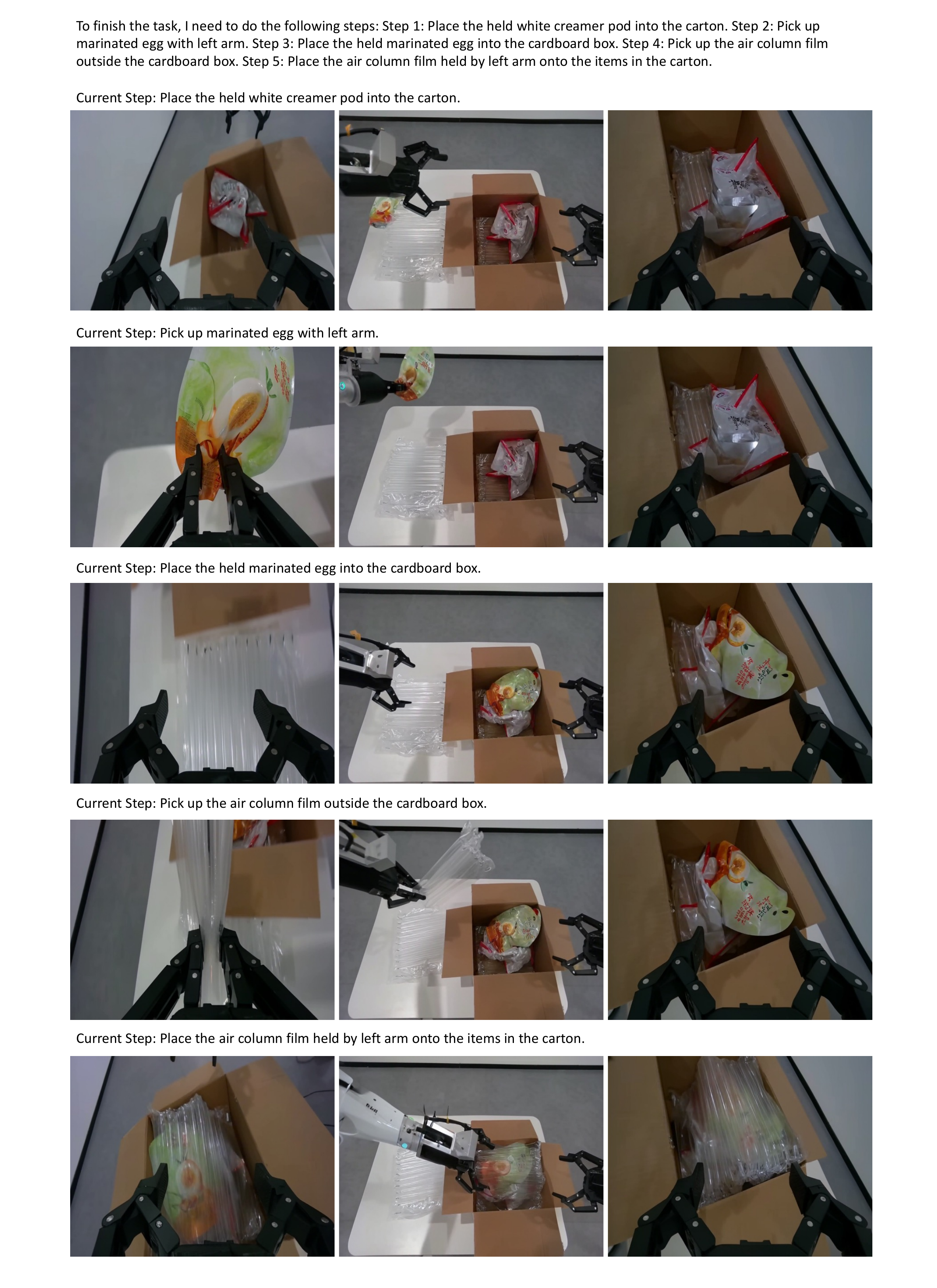}
    \caption{More interleaved generation samples produced by ScanAR posttrained U0.}
    \label{fig:interleave1}
\end{figure}

\begin{figure}[!htbp]
    \centering
    \includegraphics[width=1\linewidth]{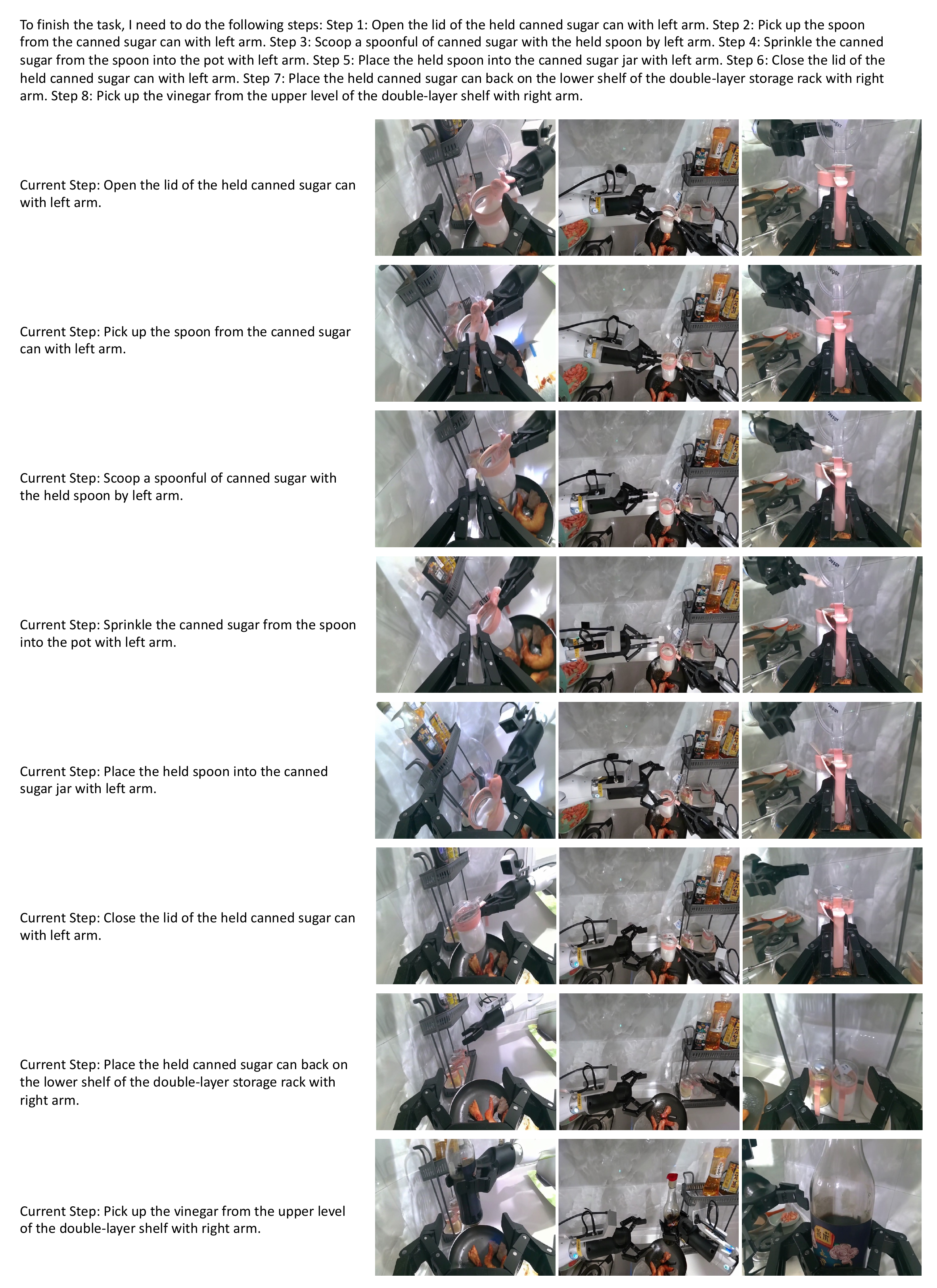}
    \caption{More interleaved generation samples produced by ScanAR posttrained U0.}
    \label{fig:interleave2}
\end{figure}

\begin{figure}[!htbp]
    \centering
    \includegraphics[width=1\linewidth]{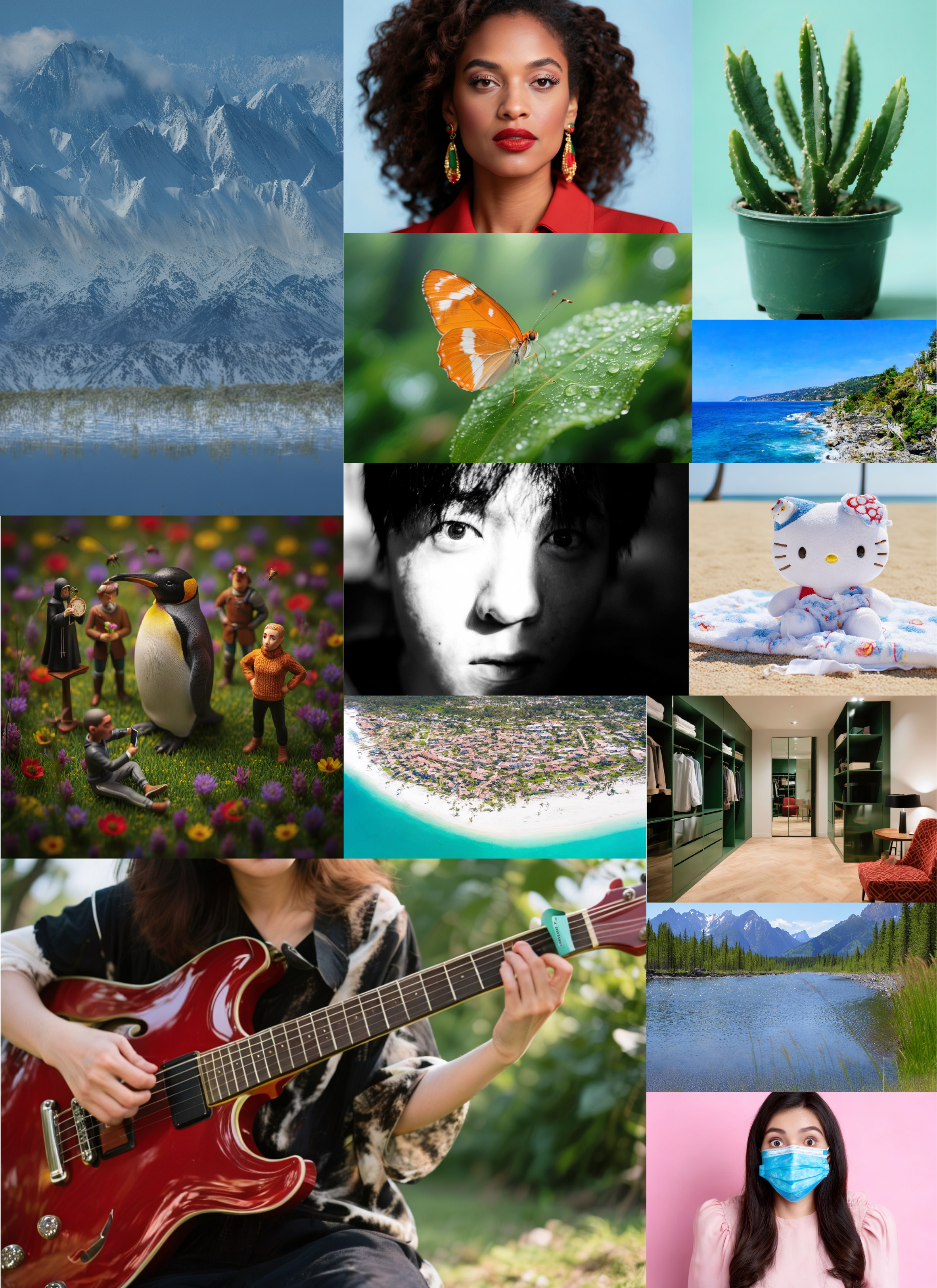}
    \caption{More text to image generation samples produced by ScanAR posttrained U0.}
    \label{fig:t2i}
\end{figure}

\begin{figure}[!htbp]
    \centering
    \includegraphics[width=1\linewidth]{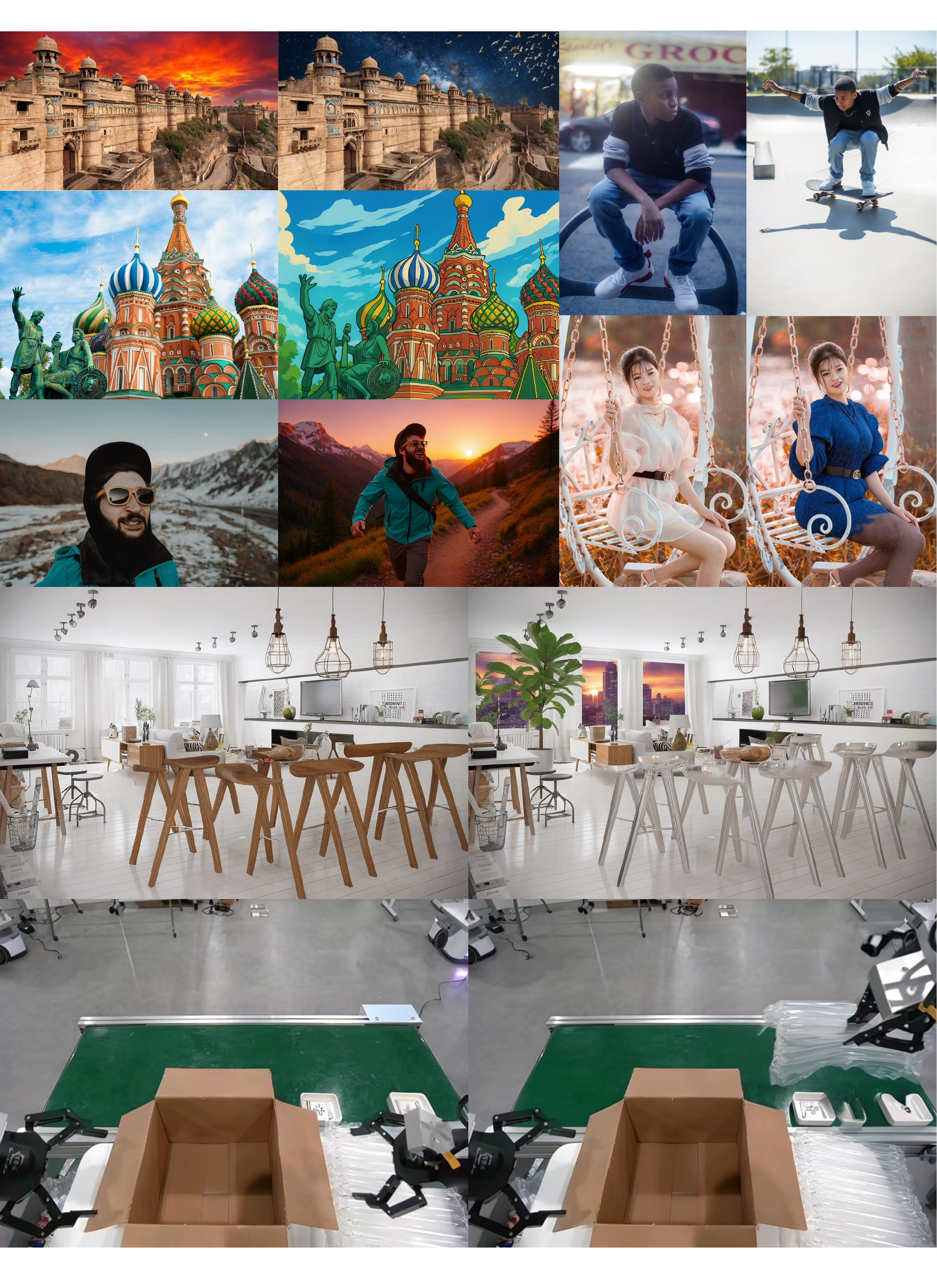}
    \caption{More image edit samples produced by ScanAR posttrained U0.}
    \label{fig:x2i}
\end{figure}

We conducted additional visualization experiments on the U0 model, including image-text interleaved sequence generation, image generation, and image editing. As shown in Figures~\ref{fig:interleave1} and \ref{fig:interleave2}, U0 retains strong image-text interleaved generation capabilities after ScanAR post-training, can freely switch between modalities during generation, preserves its unified autoregressive capability, and automatically accelerates image generation. Figure~\ref{fig:t2i} demonstrates its text to image generation capability, while Figure~\ref{fig:x2i} shows its image-editing capability. As can be seen, ScanAR accelerates image generation without compromising image quality.

\subsection{U0 Evaluation Statistics}
\label{app:u0-stats}

Table~\ref{tab:u0-stats} reports the per-sample mean and standard deviation of the GPT-5.4 scores on the 300 paired embodied-scene-transfer prompts. Every prompt produced a valid output for both models, and in none of the 600 judge explanations was a mismatch between depth maps and generated views reported, so the special rules of the evaluation prompt did not affect any score.

\begin{table}[h]
\centering
\scriptsize
\setlength{\tabcolsep}{3pt}
\caption{Per-sample GPT-5.4 scores (mean $\pm$ std) on 300 paired embodied-scene-transfer prompts.}
\label{tab:u0-stats}
\resizebox{\linewidth}{!}{%
\begin{tabular}{lccccccc}
\toprule
& Depth & Background & Foreground & Lighting & Image Quality & Multi-View & Overall \\
\midrule
U0     & $7.44\pm1.00$ & $7.60\pm1.39$ & $7.27\pm1.42$ & $7.50\pm1.63$ & $8.05\pm0.57$ & $7.52\pm1.14$ & $7.52\pm0.80$ \\
ScanAR & $7.50\pm0.91$ & $7.36\pm1.80$ & $7.21\pm1.42$ & $7.50\pm1.68$ & $7.76\pm0.71$ & $7.60\pm1.02$ & $7.49\pm0.84$ \\
\bottomrule
\end{tabular}%
}
\end{table}

\subsection{Interleaved Evaluation Metrics}
\label{app:interleave-metrics}

The 80 interleaved cases are drawn from the training split of nine robot data sources of the post-training data, and the requested number of steps is given in each task instruction. All metrics were fixed before generation. \emph{Valid output} requires that generation finishes with an end-of-sequence token and that every generated image is decoded. \emph{Number of steps} and \emph{alternation} check that the output contains the requested number of steps and that each step contains a text plan followed by the expected number of views. \emph{Image fidelity} compares each generated image with the ground-truth observation of the same step and view using DINOv2 cosine similarity, LPIPS (AlexNet), and PSNR. \emph{Scene consistency} is the DINOv2 cosine similarity between each generated image and the initial observation of the same view, and \emph{temporal consistency} the similarity between the same view in adjacent steps. \emph{Subtask-text exact match} compares the generated subtask plan of each step with the ground truth after lower-casing and removing punctuation. Generation time covers the full sequence, including text, prefill, and both CFG branches, and is divided by the number of generated images.

\subsection{LlamaGen Result}
\label{LlamaGen}
We compare ScanAR with LlamaGen, PAR, NAR, VAR, BlockDiffusion, and FlashAR on the ImageNet $256\times256$ benchmark. The results are reported in Table~\ref{tab:llamagen-short}. ScanAR is trained from scratch for 300 epochs and generates each image in 32 decoding steps.

Across all model sizes, ScanAR achieves the highest Inception Score (IS) and throughput among the compared methods. The IS improves from 244.3 to 306.9 as the model scales from B to XXL, indicating that ScanAR benefits effectively from increased model capacity. At the same time, ScanAR reduces the sequential decoding depth from 256 steps for LlamaGen to 32 steps and achieves throughputs of 669.4, 235.2, 113.4, and 65.3 images per second across the four model sizes, respectively. These results show that ScanAR substantially improves sampling efficiency while preserving strong class-conditional generation quality. The consistent gains across model sizes also demonstrate that the scan-order decoding strategy scales effectively with model capacity.

\begin{figure}[!htbp]
    \centering
    \includegraphics[width=1\linewidth]{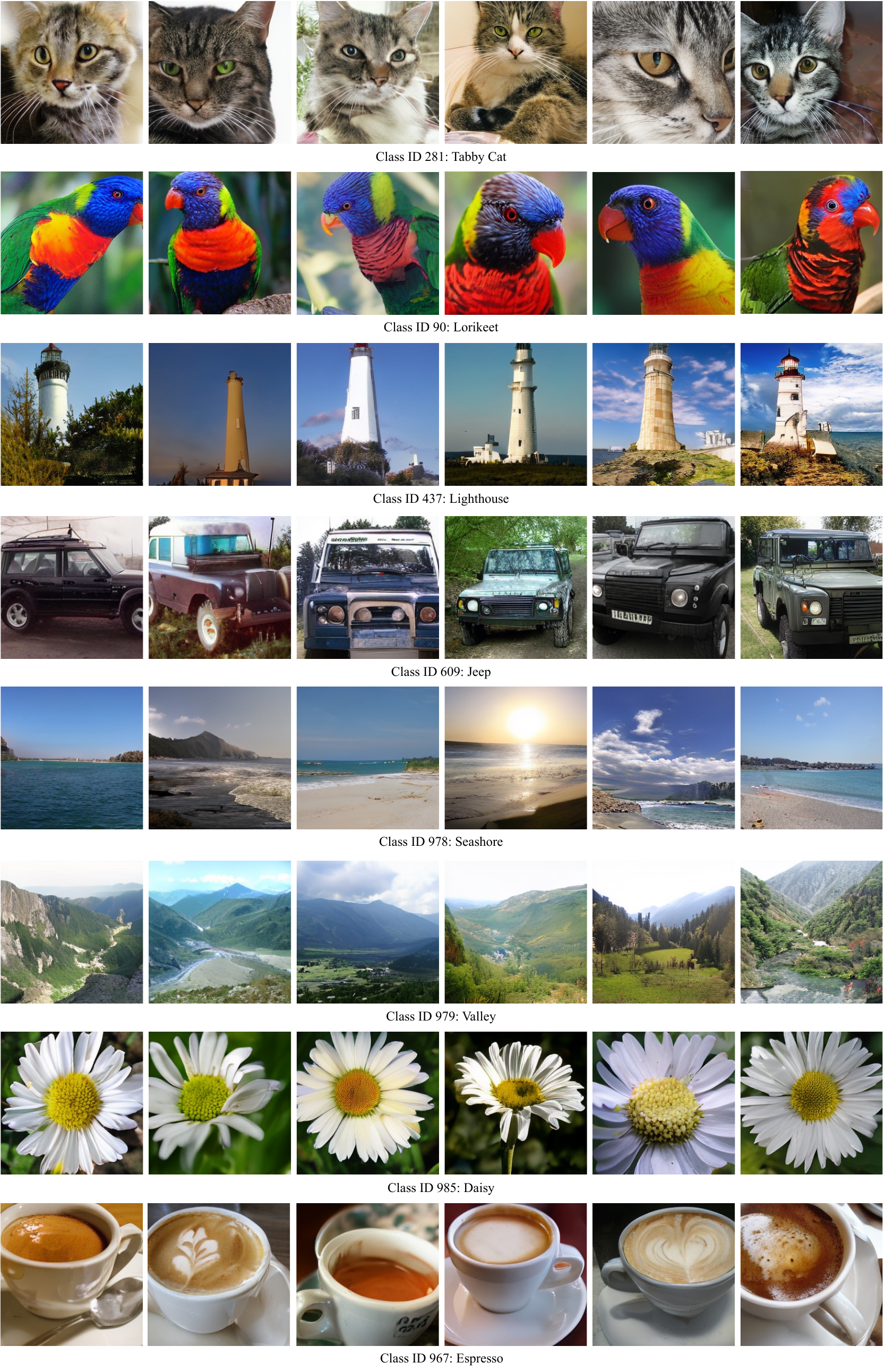}
    \caption{Class-conditional image generation samples produced by ScanAR-XXL on ImageNet \(256\times256\).}
    \label{fig:scanar-xxl}
\end{figure}

Qualitative examples of class-conditional image generation by ScanAR-XXL are shown in Figure~\ref{fig:scanar-xxl}. The generated images demonstrate strong class fidelity across diverse ImageNet categories, while exhibiting diverse appearances and rich visual details.

\subsubsection{Throughput Evaluation}
\label{app:throughput}

We evaluate the inference throughput of ScanAR on a single NVIDIA H20-3e GPU. All models are evaluated using BF16 arithmetic with PyTorch \texttt{2.9.1+cu129}. The autoregressive transformer uses the FlexAttention backend with BlockMask. We generate $256$ visual tokens arranged as a $16 \times 16$ latent grid, corresponding to $256 \times 256$ output images with a VQ downsampling factor of $16$.

The sampling configuration is fixed to classifier-free guidance (CFG) scale $2.0$, temperature $1.0$, disabled top-$k$ filtering ($k=0$), and top-$p=1.0$. Random sampling is enabled, and
the random seed is set to $0$ for reproducibility.

To reduce Python scheduling and CUDA kernel launch overhead, we use a static
CUDA Graph execution path. The graph is captured after three warm-up
generations, including cache allocation, FlexAttention compilation, and all
lazy kernel compilation. The captured graph uses a fixed batch size of
$B=512$. Before each graph replay, the class labels are copied into a
preallocated CUDA buffer outside the graph. The graph then performs the
complete ScanAR generation procedure, including the first-row causal
decoding, first-row bidirectional commit, subsequent row-wise decoding, and
token sampling.

We measure five graph replays per model. Timing is performed with CUDA events
and includes only autoregressive token generation. Checkpoint loading, graph
capture, warm-up iterations, VQ decoding, image post-processing, and disk I/O
are excluded from the timed region. Let $t_i$ denote the elapsed time of the
$i$-th timed batch. We report the generation throughput as
\begin{equation}
    \mathrm{Throughput}
    =
    \frac{B}{\frac{1}{K}\sum_{i=1}^{K} t_i},
    \qquad K=5,
\end{equation}
where the unit is generated images per second.

Table~\ref{tab:scanar_throughput} reports the measured throughput. Each result
is obtained from an independently loaded model on an otherwise idle GPU.

\begin{table}[t]
    \centering
    \caption{
        Single-GPU ScanAR generation throughput measured with a static CUDA
        Graph on an NVIDIA H20-3e. All measurements use BF16 precision,
        batch size $512$, a $16 \times 16$ visual-token grid, CFG scale
        $2.0$, temperature $1.0$, FlexAttention, three warm-up batches, and
        five timed graph replays. VQ decoding and image I/O are excluded.
    }
    \label{tab:scanar_throughput}
    \resizebox{\linewidth}{!}{%
\begin{tabular}{lccc}
        \toprule
        Model & Mean batch time (s) & Throughput (img/s)
        & Peak allocated memory (GiB) \\
        \midrule
        GPT-B  & 0.764874 & 669.39 & 12.14 \\
        GPT-L  & 2.177245 & 235.16 & 28.07 \\
        GPT-XL & 4.516144 & 113.37 & 50.52 \\
        GPT-XXL& 7.840613 &  65.30 & 79.58 \\
        \bottomrule
    \end{tabular}%
}
\end{table}

\subsection{Comparison with ZipAR}
\label{app:zipar}

Table~\ref{tab:zipar} compares ScanAR with ZipAR at different windows $w$. As the window shrinks toward ScanAR's depth, ZipAR becomes faster but its quality collapses: with 88 forward passes its FID is $3.0\times$ (B) and $1.8\times$ (L) that of ScanAR, and with 60 forward passes it degrades to 55.96 (B) and 33.90 (L). Even at $w=2$, ScanAR is $2.3\times$ (B) and $2.2\times$ (L) faster. Besides its lower depth, every ScanAR step appends one contiguous row of fixed shape, whereas each ZipAR forward writes a varying number of tokens from different rows at non-contiguous cache positions.

\begin{table}[h]
\centering
\small
\caption{Comparison with ZipAR on ImageNet $256\times256$ at the best CFG.}
\label{tab:zipar}
\resizebox{\linewidth}{!}{%
\begin{tabular}{llcrrrrr}
\toprule
Model & Method & Training & Forwards & FID$\downarrow$ & IS$\uparrow$ & P/R-F1$\uparrow$ & \shortstack{Throughput\\(img/s)} \\
\midrule
\multirow{5}{*}{GPT-B}
& Raster LlamaGen & from scratch & 256 & 5.46 & 193.6 & 0.594 & 117.9 \\
& ZipAR ($w=8$, default) & training-free & 144 & 5.69 & 192.9 & 0.568 & 158.2 \\
& ZipAR ($w=4$) & training-free & 88 & 16.59 & 89.8 & 0.574 & 227.6 \\
& ZipAR ($w=2$) & training-free & 60 & 55.96 & 28.9 & 0.464 & 288.5 \\
& \textbf{ScanAR} & from scratch & \textbf{32} & 5.59 & \textbf{244.3} & 0.568 & \textbf{669.4} \\
\midrule
\multirow{5}{*}{GPT-L}
& Raster LlamaGen & from scratch & 256 & 3.80 & 248.3 & 0.639 & 47.1 \\
& ZipAR ($w=8$, default) & training-free & 144 & 3.69 & 226.6 & 0.636 & 60.8 \\
& ZipAR ($w=4$) & training-free & 88 & 6.77 & 168.3 & 0.624 & 85.0 \\
& ZipAR ($w=2$) & training-free & 60 & 33.90 & 54.1 & 0.564 & 106.6 \\
& \textbf{ScanAR} & from scratch & \textbf{32} & 3.86 & \textbf{295.2} & 0.637 & \textbf{235.2} \\
\bottomrule
\end{tabular}%
}
\end{table}

We use the official ZipAR implementation on the official raster LlamaGen GPT-B and GPT-L checkpoints, generate 50K samples, and evaluate them with the same evaluator as ScanAR. Throughput follows Appendix~\ref{app:throughput}; to run the official decoding loop in a static CUDA graph, its per-step host index tensors are recorded during warm-up and copied from pinned memory, and its attention call is dispatched to efficient kernels instead of the math-only backend, which leaves the outputs unchanged.

\subsection{Additional Results on Emu3.5-Image-34B}
\label{app:emu-detailed-results}

\subsubsection{Training Cost and Inference Efficiency}

Table~\ref{tab:emu-efficiency} provides a detailed comparison of training
configurations and inference efficiency on Emu3.5-Image-34B. The original
Emu3.5-Image model is trained from scratch for 940K steps on 150B training
tokens. In contrast, BlockDiffusion, FlashAR, and ScanAR are post-training
methods that use 80M training tokens. ScanAR requires only 20K post-training
steps, compared with 50K steps for BlockDiffusion and FlashAR, corresponding
to a 60\% reduction in post-training steps.

\begin{table}[h]
\centering
\small
\caption{Training configurations and inference efficiency on
Emu3.5-Image-34B.}
\label{tab:emu-efficiency}
\resizebox{\linewidth}{!}{%
\begin{tabular}{l l r r r r}
\toprule
Method & Type & Steps & Data & Latency (s)$\downarrow$ & Decoding Steps \\
\midrule
Emu3.5-Image & From scratch & 940K & 150B & 64.60 & 1024 \\
BlockDiffusion & Post-training & 50K & 80M & 6.17 & 64 \\
FlashAR & Post-training & 50K & 80M & 5.68 & 63 \\
ScanAR & Post-training & 20K & 80M & 4.69 & 64 \\
\bottomrule
\end{tabular}%
}
\end{table}

The original raster-scan model requires 1024 serial decoding steps for
$512\times512$ image generation, whereas ScanAR requires only 64 steps. This
reduces the sequential decoding depth by a factor of 16. With the two CFG
branches batched in both models, the latency drops from 64.60 to 4.69 seconds,
a $13.8\times$ speedup. Compared with BlockDiffusion and
FlashAR, ScanAR also achieves lower latency, with relative reductions of
approximately $24.0\%$ and $17.4\%$. Since the accelerated methods have similar
decoding-step counts, this advantage comes from the per-step cost of ScanAR's
regular row-wise schedule.

\subsubsection{Matched Latency Measurement}
\label{app:matched-latency}

For the matched comparison in Section~\ref{sec:efficiency}, raster Emu3.5-Image
is decoded in the same inference code as ScanAR: both batch the conditional and
unconditional CFG branches into one forward pass, use a preallocated KV cache
with the same attention kernels, and run on the same H20-3e GPU with the same
eight prompts and sampler (CFG 5). We verified the raster decoder against the
reference implementation by feeding identical tokens for 32 steps: the
per-branch logits differ by 0.5\% of the logit range on average, which is
consistent with bf16 precision. A raster image takes 1{,}055 forward passes
(1{,}024 visual tokens and 31 end-of-row tokens), compared with 64 for ScanAR.
Dividing the total latency by the number of forward passes gives about 61 ms
for a single-token raster step and about 74 ms for a ScanAR row step with 32
queries, both including prefill.

\subsubsection{GenEval Per-Category Analysis}
\label{app:geneval-analysis}

Compared with the original Emu3.5-Image, ScanAR is lower on Two Objects and
Color Attribution (Table~\ref{tab:geneval-emu}). The raster model fine-tuned on
the same data is also lower on both categories, so the post-training data
lowers them for every model trained on it. On Color Attribution, ScanAR is on
par with the raster fine-tune ($68.00$ vs.\ $67.30\pm1.08$) and the best of the
three acceleration methods. On Two Objects, part of the drop is likewise due to
the data, and ScanAR remains 4.3 points below the raster fine-tune.

If tokens within a row interacted less under row-parallel decoding, relations
between objects that share image rows should suffer most. The GenEval Position
category contains both cases: ``left of / right of'' places two objects side
by side, so they share image rows, while ``above / below'' stacks them across
rows. On ScanAR's generations, same-row relations are not the weak point: they
are more accurate than cross-row relations, $84.24\%$ for left/right (184
images) versus $76.39\%$ for above/below (216 images).

\section{Implementation and Efficiency Details}
\label{app:efficiency}

\subsection{Training Implementation and Efficiency}

ScanAR augments conventional sequential sampling with parallel row-wise sampling by modifying only the attention mask and loss computation. A practical implementation challenge arises from its row-block causal mask. Replacing the standard causal mask prevents direct reuse of the optimized causal-mask execution path in FlashAttention, whereas naively materializing a dense four-dimensional (4-D) mask incurs substantial intermediate-memory and computational overhead.

\begin{table}[h]
\centering
\caption{Training efficiency of causal AR and ScanAR with different attention implementations on eight H20E GPUs (141 GB memory, TP$=4$). Step time is measured after warm-up.}
\label{tab:scanar_training_efficiency}
\small
\begin{tabular}{@{}llrr@{}}
\toprule
Sequence length & Method & Step time (s) & Peak memory (GB) \\
\midrule
\multirow{3}{*}{4K}
& Causal AR                   & 10.1 & 91.9  \\
& ScanAR with dense 4-D mask & 16.2 & 105.6 \\
& ScanAR with FlexAttention  & 10.5 & 94.5  \\
\midrule
\multirow{3}{*}{8K}
& Causal AR                   & 20.2 & 91.9  \\
& ScanAR with dense 4-D mask & --   & OOM   \\
& ScanAR with FlexAttention  & 21.3 & 94.2  \\
\bottomrule
\end{tabular}
\end{table}

We address this issue using FlexAttention, which represents the ScanAR attention pattern compactly as a \texttt{BlockMask}. The fused attention kernel processes the permitted blocks and applies online softmax without materializing the full attention-score matrix or the dense 4-D mask. This representation is well suited to ScanAR because its visible regions are composed primarily of contiguous, aligned row blocks.

Table~\ref{tab:scanar_training_efficiency} compares the resulting training efficiency with the original causal AR implementation and a ScanAR implementation using a dense 4-D mask. At a sequence length of 4K, FlexAttention reduces the ScanAR step time from 16.2 to 10.5 seconds and peak memory from 105.6 to 94.5 GB, approaching the causal AR baseline of 10.1 seconds and 91.9 GB. At 8K, the dense-mask implementation runs out of memory, whereas FlexAttention completes a step in 21.3 seconds with 94.2 GB of peak memory, compared with 20.2 seconds and 91.9 GB for causal AR.

The reported peak memory is the maximum allocated memory over the entire training step rather than the memory consumed by the attention kernel alone. In these experiments, both the causal and FlexAttention paths reached their peak allocations during other phases of the training step. Consequently, increasing the sequence length from 4K to 8K does not necessarily increase the measured peak; the small observed differences are also within normal allocator and runtime variation.

\end{document}